\documentclass[runningheads,a4paper]{llncs}

\usepackage{amssymb}
\usepackage{graphicx}
\usepackage{amsmath} 
\usepackage{subfigure}
\usepackage{caption}
\usepackage{multicol}  
\usepackage{multirow} 

\usepackage{url}
\urldef{\mailsa}\path|{solour_lfq}@sjtu.edu.cn|

\begin{document}

\mainmatter  

\title{Probabilistic Model of Object Detection Based on Convolutional Neural Network}

\titlerunning{PMOD Based on CNN}

\author{Fang-Qi Li%
\and Xu-Die Ren \and Hao-Nan Guo}
\authorrunning{F Li, X Ren, H Guo}

\institute{School of Electronic Information and Electrical Engineering\\
Shanghai Jiao Tong University\\
\mailsa\\
}

\toctitle{Lecture Notes in Computer Science}
\tocauthor{Authors' Instructions}
\maketitle

\begin{abstract}
The combination of a CNN detector and a search framework forms the basis for local object/pattern detection. To handle the waste of regional information and the defective compromise between efficiency and accuracy, this paper proposes a probabilistic model with a powerful search framework. By mapping an image into a probabilistic distribution of objects, this new model gives more informative outputs with less computation. The setting and analytic traits are elaborated in this paper, followed by a series of experiments carried out on FDDB, which show that the proposed model is sound, efficient and analytic.

\end{abstract}

\section{Introduction}
Detecting and locating local objects from images has been a challenging task in computer vision. Given some analytic, geometric patterns, this problem is handled by a branch of variants of Hough transform\cite{hough1}. But detecting and locating of abstract objects, e.g.faces, have been analytically intractable. 

So far, one of the most promising approaches is to combine convolutional neural network(CNN) with traditional methods. CNN has been taken as the most powerful mechanism for image processing, but since it was designed to solve classification problems firstly, its inflexibility in giving more informative outputs gave rise to redundancy and inefficiency of system structure in this task. The trade-off between efficiency and accuracy can be witnessed from the previous methods too. Sticking to the CNN structure as a classifier, those models served as deterministic models, which implies outputs that are less analytic.

The rest of the paper is organized as follows: Section 2 introduces traditional models' CNN structures and search frameworks. Section 3 introduces the structure and manipulation of a new probabilistic model. Experiments are presented and analyzed in section 4. Section 5 concludes the paper.

\section{Related Works}
\subsection{Previous Models}
As CNN outperforms other models in the task of recognizing abstract objects. A CNN combined with a search framework can provide a straightforward solution to this task: the search framework segments the whole image into pieces and feeds them into the CNN. However, applying CNN to every part of the image of every scale, known as exhaustive search\cite{exhaustive} is computably expensive. An alternative approach, selective search\cite{selective1} provides a heuristical segmentation of picture, which reduces complexity and accuracy. The duality of exhaustive and selective search depicts the compromise between accuracy and efficiency. As a variant of a dense artificial neural network, CNN's parameters are set to adapt to the innate trait of images, namely regional similarity, invariance under certain transformation, etc. Introducing of invariance by shared parameters increases the robustness and decreases the sensitivity at well. Thus an orthodox CNN is not well-designed to locate every trivial object from an image without a search framework. 

The combination of CNN and a selective search framework shows extraordinary result\cite{deep}. The exactitude of locating by a cascade is discussed in\cite{casade}. A recent work provides an end-to-end framework under the assumption that different layers of CNN extract features of different scales\cite{unified}. However, the problems listed below shed light on a probabilistic model.

\subsection{Problems}
Traditional detection systems have two shortcomings. First, the process of vision recognization itself is not a deterministic process, but hitherto only deterministic models are proposed. They waste the regional traits of the targeted object and the consistency of images. Second, the search framework, a trade-off between efficiency and accuracy, lacks flexibility. 

\section{Probabilistic Model of Object Detection}
\subsection{Object Distribution}
In order to use the incomplete information of a certain object, we propose a probabilistic model, together with an effective search framework and a heuristic option. As mentioned in \cite{deep}, a deformation penalty can help to locate a local object better. The existence of a object within a scope of an image can be viewed as a probabilistic distribution(as the notion of electrons in quantum physics), i.e, the cars in Fig.1 can be seen as a distribution in Fig.2. Assuming the distribution is a normal distribution, then the existence of objects within an image can be illustrated by a mixture of Gaussian:
\begin{equation}
p(\textbf{x}|\pi,\mu,\Sigma)=\sum_{n=1}^{N}\pi_{n} \frac{1}{2\pi}\frac{1}{|\Sigma_{n}|^{\frac{1}{2}}}\exp\left\{ -\frac{1}{2}(\textbf{x}-\mu_{n})^{T}\Sigma_{n}^{-1}(\textbf{x}-\mu_{n}) \right\}
\end{equation}

\begin{figure}
\begin{minipage}[t]{0.5\linewidth}
\centering
\includegraphics[width=2.4in]{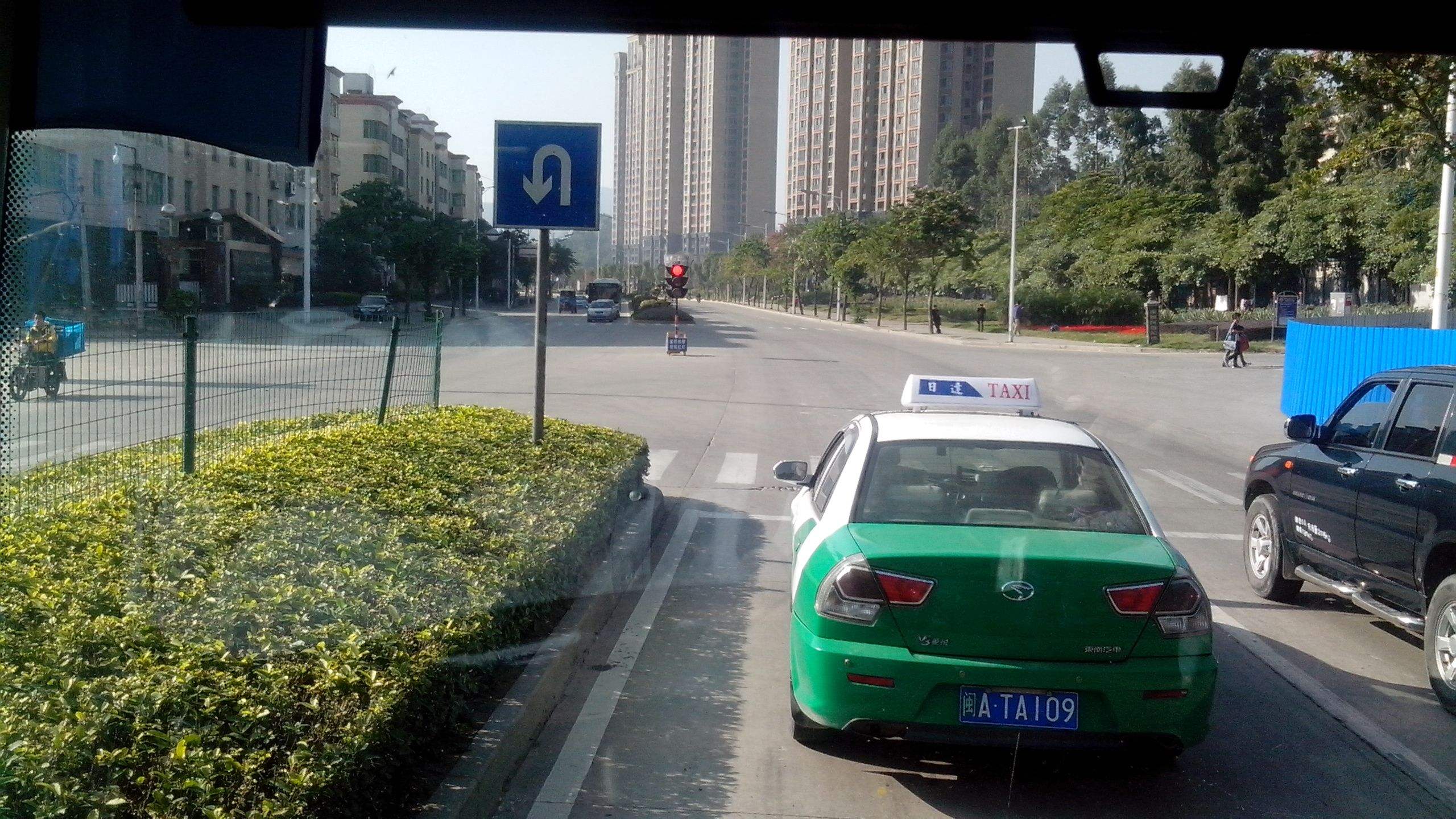}
\caption{Original Image}
\label{fig:side:a}
\end{minipage}%
\begin{minipage}[t]{0.5\linewidth}
\centering
\includegraphics[width=2.4in]{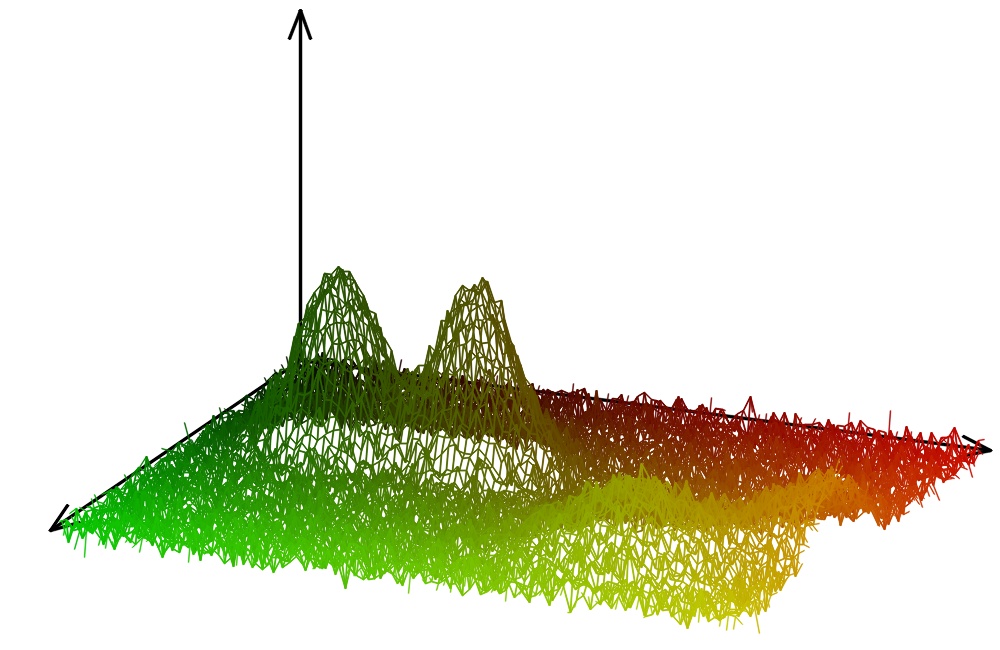}
\caption{Object Distribution of Cars}
\label{fig:side:b}
\end{minipage}
\end{figure}

Where $N$, $\mu$ and $\Sigma$ denote the number, locations and expansions, i.e.sizes of objects. We set $\pi=1$ because this object distribution is not necessarily regularized. For a probabilistic model, instead of a regular classifier, a regression task is assigned to a CNN detector. In particular, the value of the target function is the density of probabilistic distribution at the center of the receptive field. 

\subsection{Reliability and Sensitivity of the CNN detector}
We claim that such a CNN detector is only sensitive to the object of the appropriate size, i.e.an object with a suitable covariance matrix(Fig.3.(a)(b)). If an input contains too large an object, i.e.the covariance matrix has small eigenvalues or it zooms in on a part of an object, then the distribution becomes flat(Fig.3.(c)) and is likely to be submerged by noise(Fig.3.(d)). If an input contains too small an object, then the chance that it be placed in the middle of the receptive field which leads to a sharp output is relatively low. Compared with a classification detector, this detector gives outputs that are related to neighbor's, i.e.adjacent segmentations result in similar outputs. This enables the reduction of calculation, which is elaborated in section 3.4.

\begin{figure}[htbp]
\centering
\subfigure[]{
\begin{minipage}[t]{0.25\linewidth}
\centering
\includegraphics[width=1.0in]{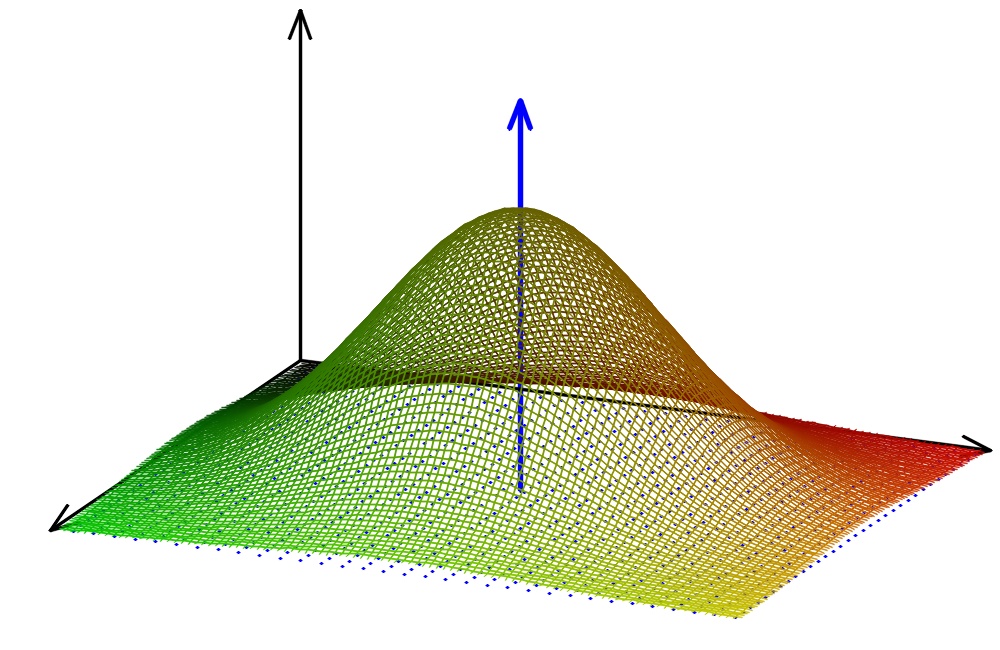}
\end{minipage}%
}%
\subfigure[]{
\begin{minipage}[t]{0.25\linewidth}
\centering
\includegraphics[width=1.0in]{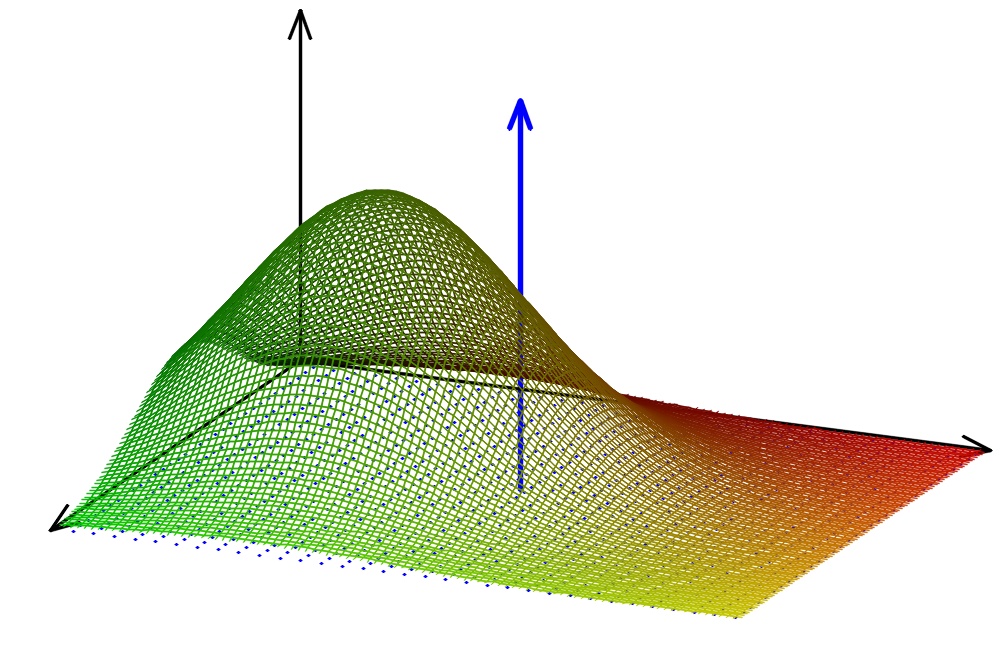}
\end{minipage}%
}%
\subfigure[]{
\begin{minipage}[t]{0.25\linewidth}
\centering
\includegraphics[width=1.0in]{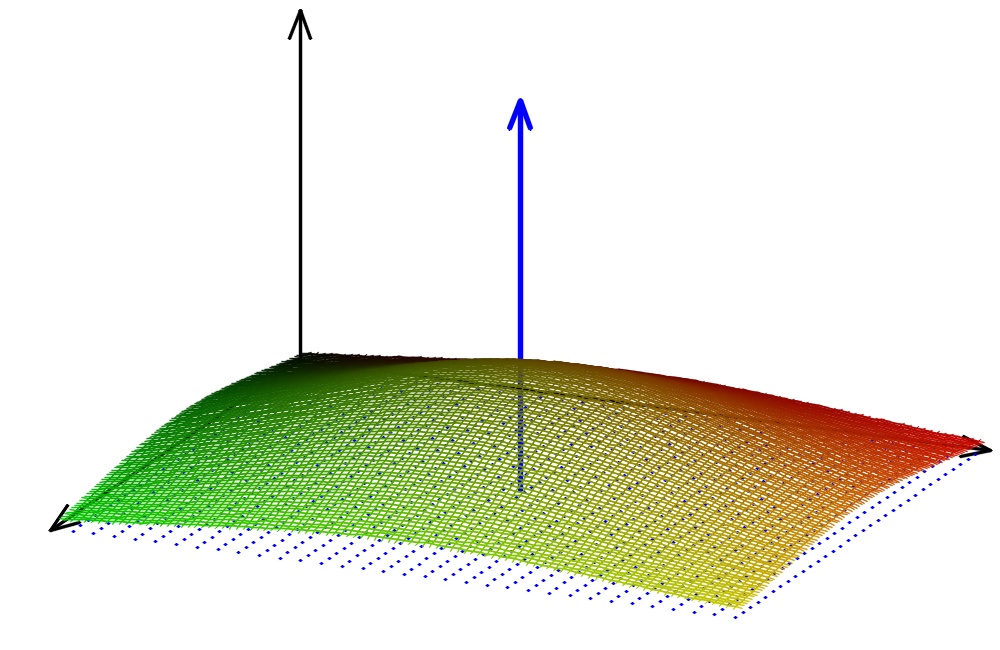}
\end{minipage}%
}%
\subfigure[]{
\begin{minipage}[t]{0.25\linewidth}
\centering
\includegraphics[width=1.0in]{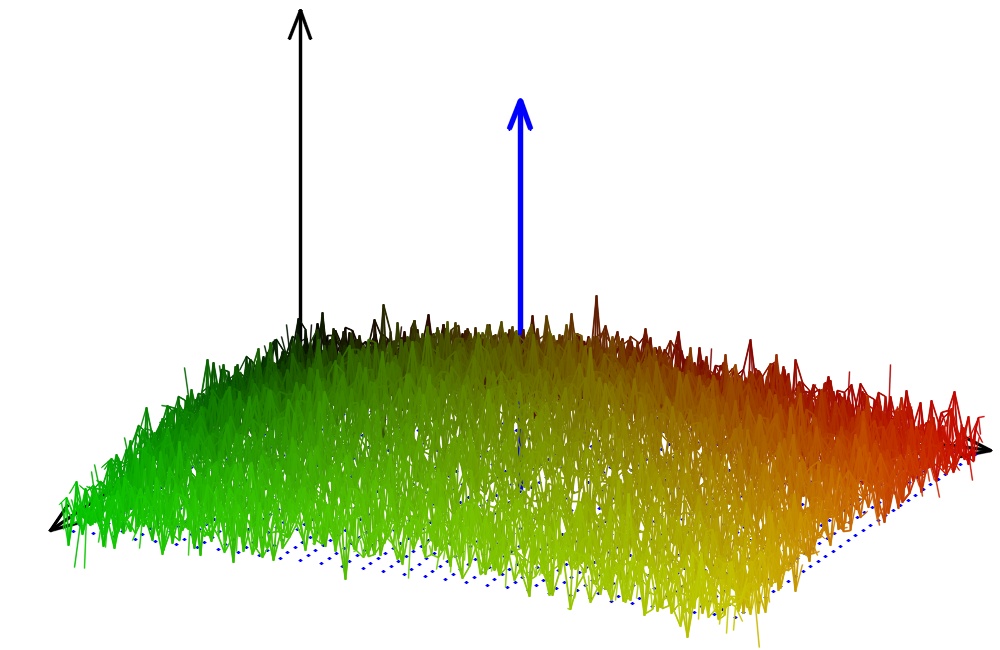}
\end{minipage}%
}%
\caption{Output of the CNN Detector in Different Circumstances}
\end{figure}

\subsection{Training of the CNN}
\subsubsection{Generated Dataset}

To train the regression function, we generate a dataset by deforming the targeted object and calculating the target values. By denoting the origin object as $\textbf{x}$, the deformation as $s$($s(\textbf{x},0)=\textbf{x}$), the target function as $f$, the deformation parameters as $\epsilon$, we have a dataset generated by $\textbf{x}$ as:
\begin{equation}
D_{\textbf{x}}=\left\{ (s(\textbf{x},\epsilon), f(s(\textbf{x},\epsilon))) \right\}
\end{equation}

The error function of this CNN structure is thus given by:
\begin{equation}
E_{0}(\textbf{w})=\frac{1}{2}\sum_{\textbf{x},\epsilon}\left\{ y(s(\textbf{x},\epsilon),\textbf{w})-f(s(\textbf{x},\epsilon)) \right\}^{2}
\end{equation}

\subsubsection{Tangent Propagation}
It can be proved that training a model with a deformed dataset equals adding extra terms onto the error function, which is known as tangent propagation\cite{tangent}. Tangent propagation was introduced to improve the robustness of traditional neural network against noise. In that sense, the error function is altered as:
\begin{equation}
E(\textbf{w})=\frac{1}{2}\sum_{n=1, \epsilon \rightarrow \textbf{0}}^{N}\left\{ y(\textbf{x}_{n},\textbf{w})-f(\textbf{x}_{n}) \right\} ^{2}+\frac{\lambda}{2} \left\{ \frac{\partial}{\partial \epsilon}y(s(\textbf{x}_{n},\epsilon),\textbf{w}) - \frac{\partial}{\partial \epsilon} f(s(\textbf{x}_{n},\epsilon)) \right\}^{2}
\end{equation}

Since:
\begin{equation}
\frac{\partial y}{\partial \epsilon} = \frac{\partial y}{\partial s} \frac{\partial s}{\partial \epsilon}
\end{equation}

The derivatives of $y$ with respect to $x$ are elements of the Jacobian matrix, which can be traced by error propagation\cite{PRML} and the derivatives of $f$ are often analytically tractable, this method played an important role in image recognition before CNN came into fashion.

In the case of our task, an alteration can be made as an analogue. To simplify the task, we assume the variance matrix to be propotional to identical matrix: 
\begin{equation}
\Sigma = \beta^{-1} I_{2}
\end{equation} 

Taking the CNN's ability of handling rotation, we further assume that the shifting takes place in only one direction. Thus $\epsilon$ consists of $\epsilon_{1}$ and $\epsilon_{2}$ that denote the change of scale and location. As the derivates are found approximately around $\epsilon \rightarrow \textbf{0}$, we can differentiate with respect to two parameters respectively.

First, a expansion in scale is denoted by $\beta \leftarrow \beta(1+\epsilon_{1})$, thus the density at the center changes from $f(s(\textbf{x},0))$ to:
\begin{equation}
f(s(\textbf{x},\epsilon_{1}))=\frac{1}{2\pi}\beta(1+\epsilon_{1})=f(\textbf{x})(1+\epsilon_{1})
\end{equation}

The derivative is inferred as:
\begin{equation}
\frac{\partial}{\partial \epsilon_{1}}f(s(\textbf{x},\epsilon_{1})) \approx \frac{f(s(\textbf{x},\epsilon_{1})) - f(s(\textbf{x},0))}{\epsilon_{1}} = f(\textbf{x})
\end{equation}


Seocnd, consider the change in location marked by $\epsilon_{2}$:
\begin{equation}
f(s(\textbf{x},\epsilon_{2}))=\frac{1}{2\pi}\beta\exp\left\{ -\frac{\epsilon_{2}^{2}}{2}\beta \right\}
\end{equation} 

Taylor expansion gives:
\begin{equation}
\frac{\partial}{\partial \epsilon_{2}} f(s(\textbf{x},\epsilon_{2})) \approx -\epsilon_{2}\beta f(\textbf{x})
\end{equation}

It is necessary to take the second order of derivative that gives:
\begin{equation}
\frac{\partial^{2}}{\partial \epsilon_{2}^{2}}f(s(\textbf{x},\epsilon_{2})) \approx -\beta f(\textbf{x})
\end{equation}

The final error function takes the form:
\begin{equation}
\begin{aligned}
E_{f}(\textbf{w}) = E_{0}(\textbf{w}) &+\frac{\lambda_{1}}{2}\sum_{\textbf{x}}\left\{ \frac{\partial}{\partial \epsilon_{1}} y(s(\textbf{x},\epsilon_{1}),\textbf{w})-f(\textbf{x}) \right\}^{2}\ \\ &+  \frac{\lambda_{2}}{2}\sum_{\textbf{x}}\left\{ \frac{\partial^{2}}{\partial \epsilon_{2}^{2}}y(s(\textbf{x},\epsilon_{2}),\textbf{w}) +\beta f(\textbf{x})\right\}^{2}
\end{aligned}
\end{equation}

Theoretically, training with a generated dataset or a modified error function leads to the same effect. So either one can be chosen to train a CNN detector.

\subsection{Search Framework}
For a better compromise between efficiency and accuracy, we propose two frameworks of search that focus on accuracy and efficiency respectively.

\begin{figure}
\begin{minipage}[t]{0.5\linewidth}
\centering
\includegraphics[width=2.4in]{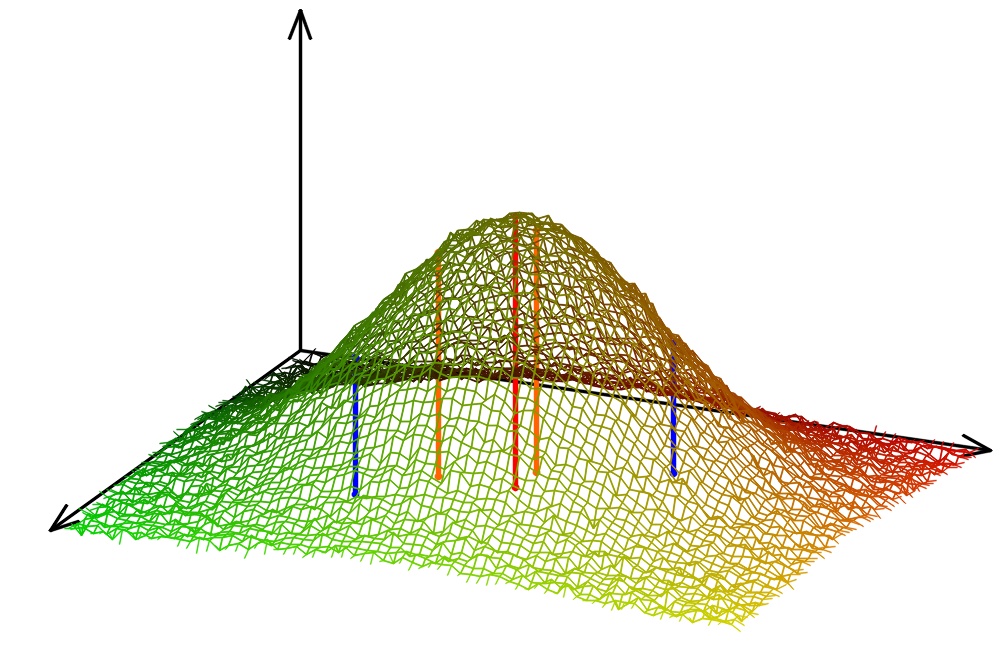}
\caption{Similirity from Adjancency}
\label{fig:side:a}
\end{minipage}%
\begin{minipage}[t]{0.5\linewidth}
\centering
\includegraphics[width=2.4in]{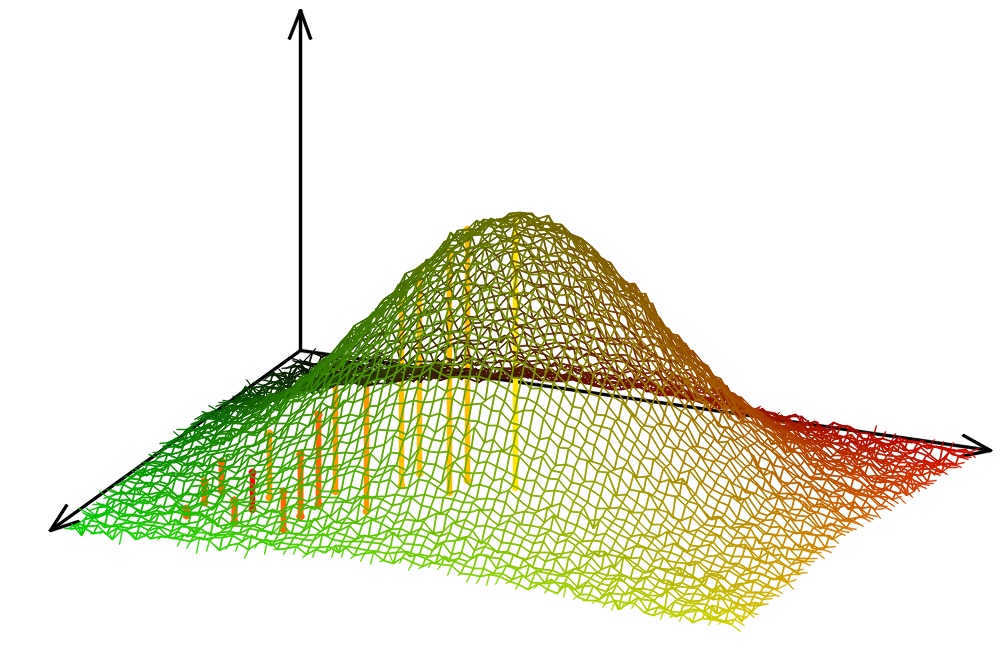}
\caption{An Ideal Search Direction}
\label{fig:side:b}
\end{minipage}
\end{figure}

\subsubsection{Exhaustive search}
Traditional exhaustive approach increases the size of the window by some pixels once, forming an arithmetic series, aiming at covering the continuous field of deformation with discrete steps of input resolution. The CNN detector based on regression task covers a quantified continuous range in deformation, namely the variance of the covariance matrix. By denoting the sensitive range of covariance for a CNN detector by $\beta^{-1} I_{2} \sim (1+\alpha)^{-1}\beta^{-1} I_{2}$(this is equivalent to setting threshohld in Fig.4), for two discrete sizes of input $s_{n}$ and $s_{n+1}$($s_{n} \le s_{n+1}$), it would be prefered if the CNN detector merges the variance seamlessly:

\begin{equation}
s_{n+1} \leq s_{n}(1+\alpha)
\end{equation}

In another word, the sizes of segmentations can be chosen as terms from a geometric series, i.e.$s_{0},s_{0}(1+\alpha), s_{0}(1+\alpha)^{2},...$, which is in accordance with intuition. Larger segmentations require a larger increase in pixels. Hence the calculation is reduced as the totality of segmentations is reduced from $O(N^{3})$ to $O(N^{2} \log N)$, where $N$ denotes the width or height of the whole image.
 

\subsubsection{Heurstical Search}
As the object distribution on the image is a scalar field and the CNN detector evaluates the scale of this field on a given point, it is possible to use the gradient information to reach the critical point. A heuristic automaton will detect the scale around itself randomly and runs in the direction with the best reward.(In Fig.5, the weakening of color depicts the process of searching). In dynamic images as videos, the automaton is capable of tracing the trajectory of a certain object. 

\section{Experiments}

\subsection{Training and Analyses}
The FDDB dataset\cite{fddb} is a labeled dataset with images containing one or more faces, the benchmark on FDDB requires recognition of local objects on a whole image, i.e.the aimed task. To generate the applicable set, we segment out 2041 faces, by shrinking/expanding(4082), shifting along axes(4000), to corners(4000), we generated a training set with 14123 images of 32 * 32. 12000 images are used to train the detector and the rest form the test set. We use a more straightforward demonstration by classifying non-confusing entries into one class, while the least square is used to train the CNN. 

Three categories of CNN are introduced: \textbf{$C_{1}$} consists of conv, pool, conv, pool, conv, dense; \textbf{$C_{2}$} consists an extra pool before dense; \textbf{$C_{3}$} is the same with \textbf{$C_{1}$} in structure with an additional term in error function. \textbf{$C_{1}$},\textbf{$C_{2}$} and \textbf{$C_{3}$} are denoted by red, green and blue in Fig.6 and Fig.7 for performace on the train set and test set after 500 times of iteration, statistics are showed in Table 1.

\begin{figure}
\setlength{\abovecaptionskip}{10pt}
\setlength{\belowcaptionskip}{0pt}
\begin{minipage}[t]{0.5\linewidth}
\centering
\includegraphics[width=1.6in]{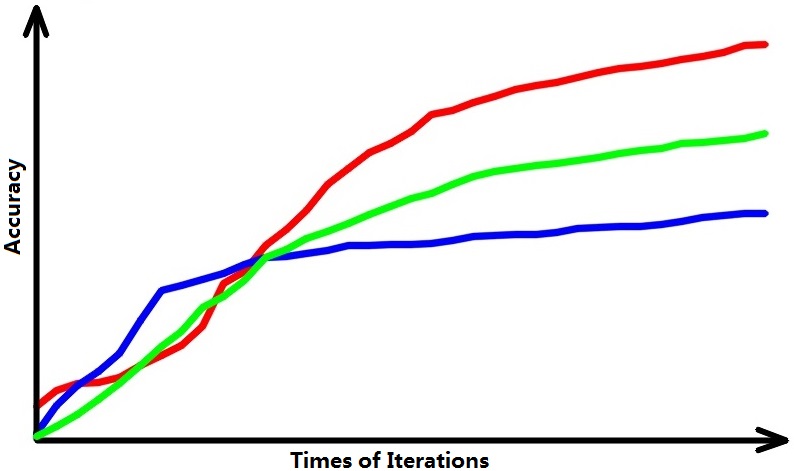}
\caption{Accuracy : 95.7\%,84.9\%,73.8\% }
\label{fig:side:a}
\end{minipage}%
\begin{minipage}[t]{0.5\linewidth}
\centering
\includegraphics[width=1.6in]{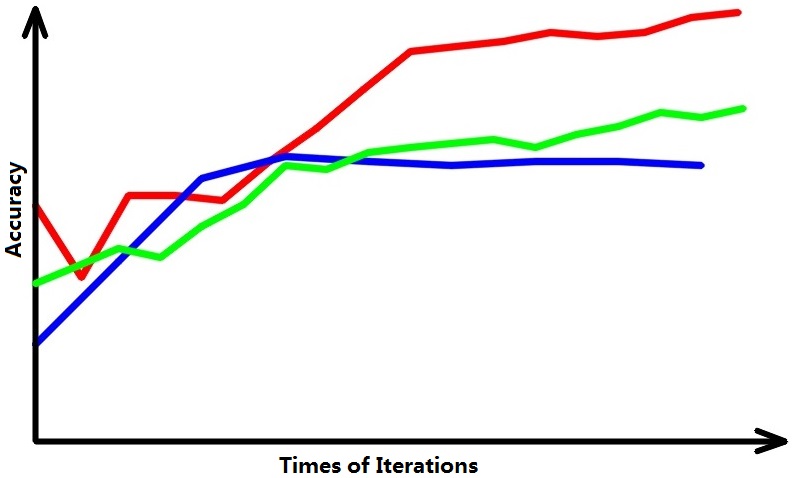}
\caption{Accuracy : 84.5\%, 76.5\%, 71.5\% }
\label{fig:side:b}
\end{minipage}
\end{figure}

\renewcommand{\arraystretch}{1.5}
\begin{table}[!thpb]  
\setlength{\abovecaptionskip}{0pt}
  \centering  
  \fontsize{10}{10}\selectfont  
  \caption{Performance of Models in Datasets w.r.t Iteration Times}  
  \label{tab:performance_comparison}  
    \begin{tabular}{|c|c|c|c|c|c|c|}  
    \hline  
    \multirow{2}{*}{Accuracy(\%)}&  
    \multicolumn{2}{c|}{100 times}&\multicolumn{2}{c|}{300 times}&\multicolumn{2}{c|}{500 times}\cr\cline{2-7}  
    &Train&Test&Train&Test&Train&Test\cr  
    \hline  
    \hline  
    \textbf{$C_{1}$}&59.7&65.6&87.5&81.0&95.7&84.5\cr\hline  
    \textbf{$C_{2}$}&61.8&62.3&80.1&74.5&84.9&76.5\cr\hline  
    \textbf{$C_{3}$}&67.3&67.3&73.4&71.0&73.8&71.5\cr\hline   
    \hline  
    \end{tabular}  
\end{table}

An extra pooling layer reduces the accuracy as it causes extra loss of information. Introducing extra terms in error function increases the convergence speed at the initial phases, but it fails to react to sharp variances in the data set. But this method does not need a generated input (\textbf{$C_{3}$} trained on 2041 original images uses the other 12082 images as test set, leads to the ordinary performance in both Fig.6 and Fig.7.)

\subsection{Comparison}
\begin{figure}[!htbp]
\centering
\includegraphics[width=4.0in]{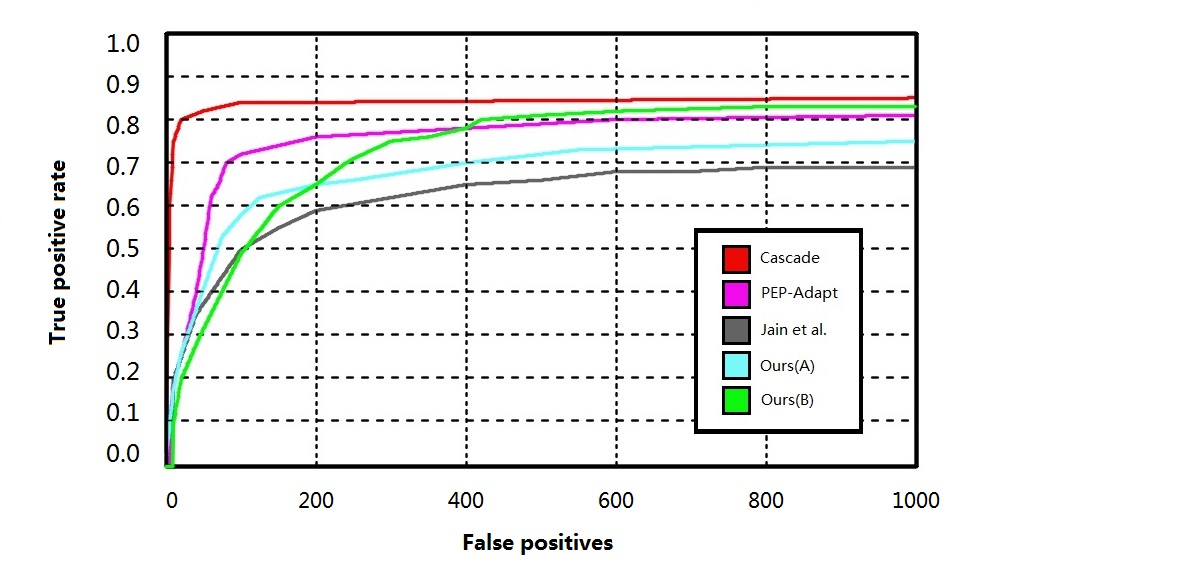}
\caption{Comparation on the FDDB dataset with state-of-the-art Models.}
\label{fig:example}
\end{figure}

Using the discontinuous score on FDDB(where the size is taken according to the covariance matrix), we compare our models with state-of-the-art models including PEP-Adapt\cite{pep}, Jain \emph{et al.}\cite{jain} and the CNN Cascade. Model A uses naive exhaustive search framework. Model B is added with an additional evidence justifier, for each assumed object, calculations around it are done to see whether the object is mere noise. The results are showed in Fig.8.

Our model is second to only few models, e.g. the cascade for face detection. However, our model is much cheaper in computing complexity with only one CNN. Since the way we handle the detector's outputs is quite bald, it is possible that improvement in handling the output of the probabilistic detector, e.g.a Bayesian approach, can result in a better performance.

\section{Conclusion}
In this work, we propose a probabilistic model for object detecting. A CNN is assigned a regression task, which makes use of regional information for a more sophisticated output. It is elaborated that the deformation of objects, which has been an obstacle for object recognition, can become informative evidence for detecting and locating. This probabilistic model is open to more delicate processors, which makes better effects on this task attainable.

\section*{Acknowledgement}
This research work is funded by the National Key Research and Development Project of China (2016YFB0801003), Key Laboratory for Shanghai Integrated Information Security Management Technology Research，Science and Technology Project of State Grid Corporation of China (SGCC)

\end{document}